\documentclass[runningheads]{llncs}

\usepackage{tikz}
\usepackage{pgfplots}
\usepackage{pgfplotstable}
\usepackage{graphicx}
\usepackage{amsmath}
\usepackage{amsfonts}
\usepackage{caption}
\usepackage{subcaption}
\usepackage[hidelinks]{hyperref}
\usepackage{todonotes}
\usepackage{xspace}
\usepackage{cite}

\renewcommand{\todo}[1]{}

\def\figref#1{Fig.~\ref{#1}}
\def\subfigref#1{(\subref{#1})}
\newcommand\scalemath[2]{\scalebox{#1}{\mbox{\ensuremath{\displaystyle #2}}}}
\newcommand{\changed}[1]{\textcolor{black}{#1}}

\DeclareMathOperator{\dist}{dist}
\DeclareMathOperator{\evid}{evid}
\DeclareMathOperator{\curv}{curv}

\usepackage{xcolor}
\definecolor{color_1}{HTML}{82CAFC}
\definecolor{color_2}{HTML}{C875C4}
\definecolor{color_3}{HTML}{580F41}
\definecolor{color_4}{HTML}{2E294E}
\definecolor{color_5}{HTML}{C2BBF0}

\definecolor{color_bright}{HTML}{FFFFFF}
\definecolor{color_lightgrey}{HTML}{EEEEEE}
\definecolor{color_grey}{HTML}{666666}
\definecolor{color_dark}{HTML}{3D3D35}
\colorlet{color_highlight}{color_2}

\usetikzlibrary{arrows}
\usetikzlibrary{arrows.meta}
\usetikzlibrary{backgrounds}
\usetikzlibrary{calc}
\usetikzlibrary{chains}
\usetikzlibrary{decorations.markings}
\usetikzlibrary{decorations.pathmorphing}
\usetikzlibrary{fit}
\usetikzlibrary{matrix}
\usetikzlibrary{mindmap}
\usetikzlibrary{positioning}
\usetikzlibrary{shadows}
\usetikzlibrary{shapes}
\usetikzlibrary{trees}

\tikzstyle{image}=
  [rectangle,draw,fill=white,inner sep=2pt]

\tikzstyle{candidate}=
  [circle,draw,fill=white,minimum height=4.5mm,inner sep=0]
\tikzstyle{candidate_edge}=
  [draw=color_grey]
\tikzstyle{candidate_subset}=
  [line width=6mm,cap=round,line join=round,opacity=0.5]

\tikzstyle{data}=
  [rectangle,rounded corners,draw,fill=gray!10!white]
\tikzstyle{method}=
  [rectangle,fill=funkey_color,text width=2cm,align=center]
\tikzstyle{arrow}=
  [->,very thick]
\tikzstyle{process}=
  [single arrow,shading=axis,left color=orange!50!white, right color=orange!40!white,draw=orange!50!white!80!black,inner sep=2pt,minimum height=8mm,single arrow head extend=2mm]

\newcommand{\getzoomfactor}{%
\pgfgettransformentries{\myxscale}{\@tempa}{\@tempa}{\myyscale}{\@tempa}{\@tempa}
\gdef\zoomfactor{\myxscale}
}

\pgfplotsset{compat=1.13}
\usepgfplotslibrary{statistics}

\tikzstyle{filledbox}=[
  fill,
  fill opacity=0.3,
  mark=o,
  mark size=0.5,
  solid]

\begin{document}

\title{Microtubule Tracking in Electron Microscopy Volumes}
\titlerunning{Microtubule Tracking in Electron Microscopy Volumes}
\author{Nils Eckstein\textsuperscript{1,2,*}, 
        Julia Buhmann\textsuperscript{1,2}, 
        Matthew Cook\textsuperscript{2}, 
        Jan Funke\textsuperscript{1}}
\authorrunning{Eckstein et al.}
\institute{\changed{\textsuperscript{1}HHMI Janelia Research Campus, Ashburn, USA,\\
           \textsuperscript{2}Institute of Neuroinformatics UZH/ETHZ, Zurich, Switzerland,\\
           \textsuperscript{*}Corresponding author: ecksteinn@janelia.hhmi.org}}
\maketitle

\begin{abstract}
  %
  We present a method for microtubule tracking in electron microscopy volumes.
  %
  %
  Our method first identifies a sparse set of voxels that likely belong to microtubules. 
  Similar to prior work, we then enumerate potential edges between
  these voxels, which we represent in a candidate graph.
  Tracks of microtubules are found by selecting nodes and edges in the
  candidate graph by solving a constrained optimization problem incorporating
  biological priors on microtubule structure.
  %
  %
  %
  For this, we present a novel integer linear programming formulation, which
  results in speed-ups of three orders of magnitude and an increase of 53\%
  in accuracy compared to prior art (evaluated on three $1.2\times4\times4\mu m$ volumes 
  of \textit{Drosophila} neural tissue).
  We also propose a scheme to solve the optimization problem in a
  block-wise fashion, which allows distributed tracking and is necessary to
  process very large electron microscopy volumes.
  %
  %
  Finally, we release 
  a benchmark dataset for microtubule tracking, here used for training, 
  testing and validation, consisting of eight 30 x 1000 x 1000 voxel blocks ($1.2\times4\times4\mu m$) 
    of densely annotated microtubules in the CREMI data set (\changed{\url{https://github.com/nilsec/micron}}).
\end{abstract}

\section{Introduction}

\begin{figure}[t]
    \centering
    \includegraphics[width=1.0\linewidth]{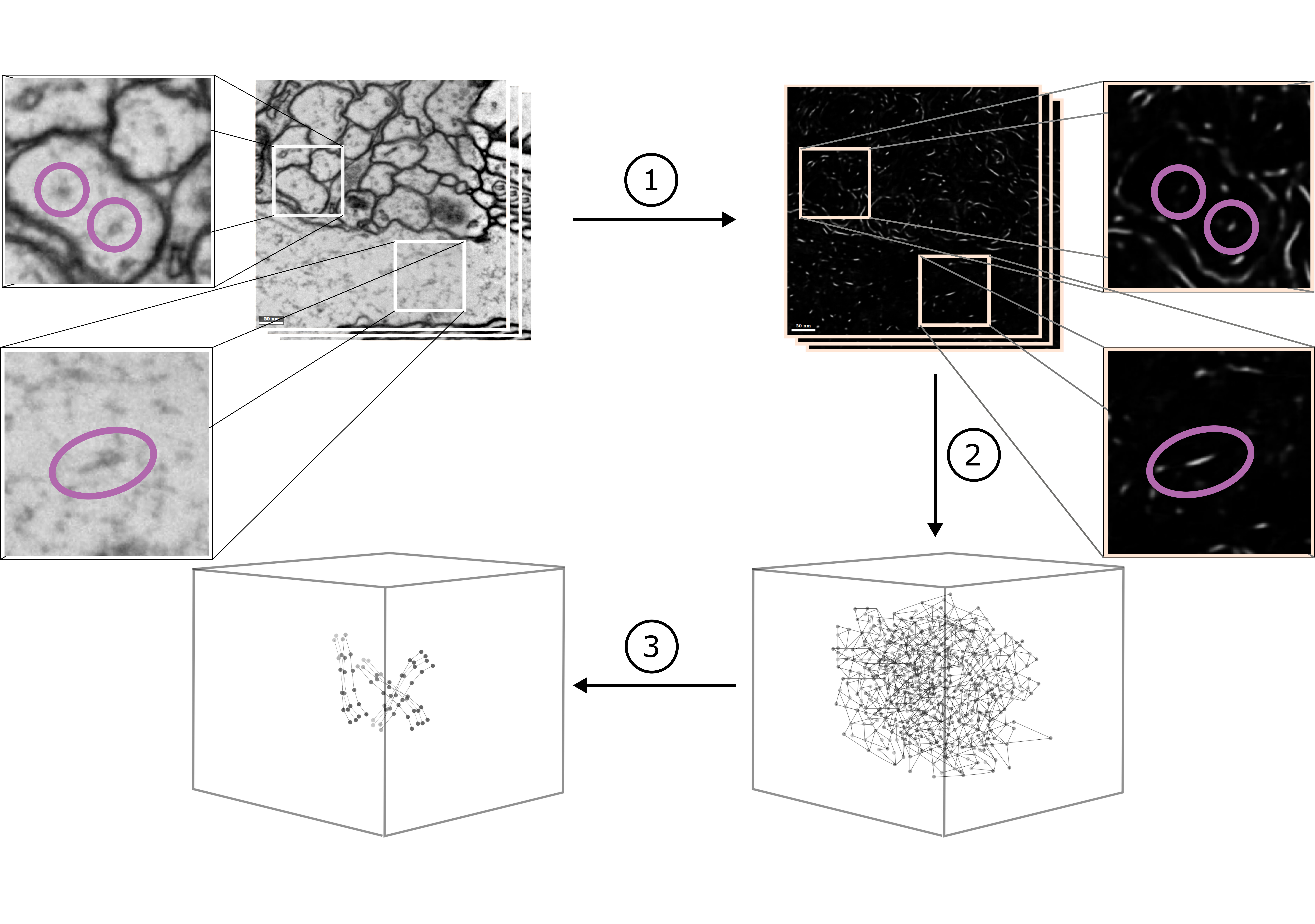}
    \caption{Overview of the proposed method. \textbf{1.} Microtubule scores
    are predicted via a 3D UNet~\cite{ronneberger2015u}. \changed{Inlets show two microtubules that run perpendicular to the imaging plane (appearing as circles) and one that deviates from a 90 degree angle of incidence (appearing as a line segment). The corresponding (noisy) microtubule scores show the necessity of post processing.} \textbf{2.} Candidate
    microtubule segments are extracted and represented as vertices in a 3D
    graph, where vertices are connected within a threshold distance.
    \textbf{3.} Final microtubule trajectories are found by solving a
    constrained optimization problem.}
    \label{fig:method:overview}
\end{figure}

%
Microtubules are part of the cytoskeleton of a cell and crucial for a variety
of cellular processes such as structural integrity and intracellular transport
of cargo~\cite{mt_review_0}.
  They are of particular interest for the connectomics community, as they
  directly follow the morphology of neurons.
  Tracking of microtubules therefore provides additional structural information
  that can potentially be leveraged for guided proof-reading of neuron
  segmentation \changed{and aid in the identification of neural subcompartments such as backbones and
  twigs}~\cite{schneider2016quantitative}.

%
Manual tracking of microtubules faces the same limitations as neuron
segmentation and synapse annotations.
  The resolution needed to discern individual structures of interest like
  neural arbors, synapses, and microtubules can only be achieved with high
  resolution electron microscopy (EM), which results in large datasets (several hundred terabytes) even for
  small model organisms like \emph{Drosophila
  melanogaster}~\cite{zheng2018complete}.
  %
  %
  With datasets of these sizes, a purely manual analysis becomes impractical.
  Consequently, the field of connectomics sparked a surge of automatic methods
  to segment
  neurons (for recent advances \changed{see}~\cite{mala,januszewski2018high,Lee2019,Lee2017superhuman}),
  annotate \changed{synapses}~\cite{synapse_julia,synapses_02,synapses_03,Buhmann2019,heinrich2018synaptic,huang2018fully,mitos3}, and identify other structures of
  biological relevance such as microtubules~\cite{Buhmann2016} or \changed{mitochondria}~\cite{mitos,mitos2,mitos3}.
%
Large scale automatic reconstruction of microtubules is a particularly
challenging problem.
  With an outer diameter of $24~nm$, microtubules are close to the resolution limit
  of serial section EM\footnote{Resolution is around $4\times 4 \times 40~nm$
  for ssTEM, and $8\times 8 \times 8~nm$ for FIB-SEM~\cite{FIB25}.}.
  Especially in anisotropic EM volumes, the appearance of microtubules changes
  drastically depending on their angle of incidence to the imaging plane.
  Furthermore, they are often locally indistinguishable from other cell
  organelles (like endoplasmic reticulum) or noise.\todo{Are synaptic vesicles a problem?}
  %

%
%
Our method for microtubule tracking is based on the formulation proposed
in~\cite{Buhmann2016}, with significant improvements in terms of efficiency and accuracy.
  Similar to~\cite{Buhmann2016}, we first predict a score for each voxel to be
  part of a microtubule. We then identify promising candidate points and
  possible links between them in a candidate graph as nodes and edges. Finally,
  we solve a constraint optimization problem incorporating biological priors to
  find a subset of edges that constitute microtubule tracks (for an overview
  see~\figref{fig:method:overview}).\\
  Our four main contributions are as follows:
  1. We propose a new integer linear program (ILP) formulation, which decreases
  the time needed to solve the constraint optimization by several orders of
  magnitude.
  2. We devise a scheme to solve the resulting optimization problem in a
  block-wise fashion in linear time, and thus are able to process real-world 
  sized volumes.
  3. Our formulation allows tracking of microtubules in arbitrary orientations
  in anisotropic volumes by introducing a non-maxima suppression (NMS) based 
  candidate extraction method.
  4. We improve the voxel-level classifier by training a 3D
  UNet~\cite{ronneberger2015u,mala} on skeleton annotations, leading to more
  accurate microtubule scores.

  We evaluate our method on a new benchmark comprising $153.6~\mu m^3$ of densely
  traced microtubules, demonstrating a 53\% increase in accuracy \changed{($0.517 \rightarrow 0.789$ F1 score)} compared to 
  the prior state of the art.
  Source code and datasets are publicly available \changed{at}~\url{https://github.com/nilsec/micron}.

\section{Method}
\subsection{\changed{Predictions}}
Starting from the raw EM input data, we train a 3D UNet~\cite{ronneberger2015u} to predict a microtubule score $m \in [0,1]$ for each voxel. 
We generate microtubule scores for training from manually annotated skeletons by interpolating between skeleton markers on a voxel grid followed by Gaussian smoothing. 
In addition, we train the network to predict spatial gradients of the microtubule score up to second order. 
This is motivated by the idea that the spatial gradient encodes the local shape of a predicted object. 
Since microtubule segments have locally line-like shapes this auxiliary task potentially regularises microtubule score predictions.
%
\subsection{\changed{Candidate Extraction}}
Given the predicted microtubule score we perform candidate extraction via two NMS passes, 
to guarantee that two successive candidates of a single microtubule
track are not farther apart than the distance threshold $\theta_d$ we will
use to connect two candidates with each other. 
In a first pass, we perform NMS and thresholding with a stride equal to the NMS window size, guaranteeing at least one candidate per NMS window if the maximum is above the threshold. 
This strategy is problematic if the local maximum lies on the boarder or corner of a NMS window as this produces multiple, in the worst case eight, candidates that are direct neighbors of each other.
We remove this redundancy by performing a second NMS pass on the already extracted maxima, providing us with the final set of microtubule segment candidate detections $C$.
\begin{figure}[t]
\end{figure}

\begin{figure}[t]
  \tikzstyle{ilp}=
    [matrix of math nodes,nodes in empty cells,anchor=north,draw=black]
  \begin{subfigure}[t]{\textwidth}
    \begin{tikzpicture}[scale=0.5]

  \foreach \i in {1, ..., 3}{

    \begin{scope}[xshift=\i*5 cm]

      \node[candidate] (v) at (0, 0) {$j$};
      \node[candidate] (n1) at (-1, -1) {};
      \node[candidate] (n2) at (-1, 1) {};
      \node[candidate] (n3) at (1, 1) {};
      \node[candidate] (n4) at (1, -1) {$k$};

      \draw[candidate_edge] (v) -- (n1);
      \draw[candidate_edge] (v) -- (n2);
      \draw[candidate_edge] (v) -- (n3);
      \draw[candidate_edge] (v) -- (n4);

      \begin{pgfonlayer}{background}
        \draw[color=color_1,candidate_subset] (n4.center) -- (v.center) -- (n\i.center);
      \end{pgfonlayer}
      \node at (n\i.center) {$i$};

    \end{scope}

  }

  \begin{scope}[xshift=4*5 cm]

    \node[candidate] (v) at (0, 0) {$j$};
    \node[candidate] (n1) at (-1, -1) {$k$};
    \node[candidate] (n2) at (-1, 1) {$i$};
    \node[candidate] (n3) at (1, 1) {};
    \node[candidate] (n4) at (1, -1) {};

    \draw[candidate_edge] (v) -- (n1);
    \draw[candidate_edge] (v) -- (n2);
    \draw[candidate_edge] (v) -- (n3);
    \draw[candidate_edge] (v) -- (n4);

    \begin{pgfonlayer}{background}
      \draw[color=color_1,candidate_subset] (n1.center) -- (v.center) -- (n2.center);
    \end{pgfonlayer}

  \end{scope}

  \node at (2.5+5, 0) {$+$};
  \node at (2.5+10, 0) {$+$};
  \node at (2.5+15, 0) {$+$};
  \node at (2.5+20, 0) {$+$};
  \node at (2.5+21.5, 0) {$\cdots$};
  \node at (2.5+23, 0) {$\leq$};
  \node at (2.5+24, 0) {$1$};

\end{tikzpicture}
    \begin{tikzpicture}[scale=0.5]

  \foreach \i in {1, ..., 2}{

    \begin{scope}[xshift=\i*6 cm]

      \node[candidate] (u) at (0, 0) {$i$};
      \node[candidate] (v) at (1.5, 0) {$j$};
      \node[candidate] (n1) at (-1, -1) {};
      \node[candidate] (n2) at (-1, 1) {};
      \node[candidate] (n3) at (2.5, 1) {};
      \node[candidate] (n4) at (2.5, -1) {};

      \draw[candidate_edge] (u) -- (v);
      \draw[candidate_edge] (u) -- (n1);
      \draw[candidate_edge] (u) -- (n2);
      \draw[candidate_edge] (v) -- (n3);
      \draw[candidate_edge] (v) -- (n4);

      \begin{pgfonlayer}{background}
        \draw[color=color_1,candidate_subset] (n\i.center) -- (u.center) -- (v.center);
      \end{pgfonlayer}
      \node at (n\i.center) {$k$};

    \end{scope}

  }

  \foreach \i in {3, ..., 4}{

    \begin{scope}[xshift=\i*6 cm]

      \node[candidate] (u) at (0, 0) {$i$};
      \node[candidate] (v) at (1.5, 0) {$j$};
      \node[candidate] (n1) at (-1, -1) {};
      \node[candidate] (n2) at (-1, 1) {};
      \node[candidate] (n3) at (2.5, 1) {};
      \node[candidate] (n4) at (2.5, -1) {};

      \draw[candidate_edge] (u) -- (v);
      \draw[candidate_edge] (u) -- (n1);
      \draw[candidate_edge] (u) -- (n2);
      \draw[candidate_edge] (v) -- (n3);
      \draw[candidate_edge] (v) -- (n4);

      \begin{pgfonlayer}{background}
        \draw[color=color_1,candidate_subset] (u.center) -- (v.center) -- (n\i.center);
      \end{pgfonlayer}
      \node at (n\i.center) {$k$};

    \end{scope}

  }

  \node at (3.75+6, 0) {$+$};
  \node at (3.75+12, 0) {$=$};
  \node at (3.75+18, 0) {$+$};

\end{tikzpicture}
    \caption{Consistency constraints (top row) and no-branch constraints (bottom row).}
    \label{ref:method:ilp:constraints}
  \end{subfigure}

  \begin{subfigure}[t]{0.5\textwidth}
    \begin{tikzpicture}[every node/.style={scale=0.75,transform shape}]
      \matrix[ilp] (old_ilp) {
          \min_{I_{i,j,k}}            &\scalemath{0.8}{\displaystyle \sum_{i\in V}c_i\cdot I_{i} + \sum_{(i,j)\in E} c_{i,j}\cdot I_{i,j} + \sum_{(i,j,k)\in T}c_{i,j,k}\cdot} &\scalemath{0.8}{I_{i,j,k}}  \\
        \text{s.t.}               & & \\
        \forall\; (i, j, k) \in T:& I_i, I_{i,j}, I_{i,j,k} \in \{0,1\} & \\
        \forall\; i \in V:        &
                                  \displaystyle
                                  2 I_i - \sum_{(i,j) \in E} I_{i,j} &
                                  = 0 \\
        \forall\; (i, j) \in E:   &
                                  2 I_{i,j} - I_i - I_j &
                                  \leq 0 \\
        \forall\; (i, j, k) \in T:&
                                  2 I_{i,j,k} - I_{i,j} - I_{j,k} &
                                  \leq 0 \\
                                  &
                                  - I_{i,j,k} + I_{i,j} + I_{j,k} &
                                  \leq 1 \\
      };
    \end{tikzpicture}
    \caption{ILP following~\cite{Buhmann2016}.}
    \label{ref:method:ilp:old}
  \end{subfigure}
  \hfill
  \begin{subfigure}[t]{0.5\textwidth}
    \hspace{3mm}
    \begin{tikzpicture}[every node/.style={scale=0.75,transform shape}]
      \matrix[ilp] (new_ilp) {
          \min_{I_{i,j,k}}            &\displaystyle \sum_{(i,j,k)\in T}c_{i,j,k}\cdot I_{i,j,k} & \\
        \text{s.t.}               & & \\
        \forall\; (i,j,k) \in T:  & I_{i,j,k} \in \{0,1\} & \\
        \forall\; j \in V:        &
                                  \displaystyle
                                  \sum_{(i,j,k)\in T} I_{i,j,k} & \leq 1 \\
        \forall\; (i, j) \in E:   &
                                  \displaystyle
                                  \sum_{(k,i,j)\in T}
                                  I_{k,i,j} -
                                  \sum_{(i,j,k)\in T}
                                  I_{i,j,k}
                                  & = 0 \\
      };
      \node at (0,-3.3) {};
    \end{tikzpicture}
    \caption{Reformulated ILP on triplet indicators.}
    \label{ref:method:ilp:new}
  \end{subfigure}
  \caption{Constraint optimization on the candidate graph. We formulate an 
  ILP on binary triplet indicators, which encode the
  joint selection of two incident candidate edges. The constraints shown
  in~\subfigref{ref:method:ilp:constraints} ensure that found tracks are not
  crossing or splitting. Although mathematically equivalent to the formulation
  in~\subfigref{ref:method:ilp:old}, our formulation~\subfigref{ref:method:ilp:new}
  is orders of magnitudes more efficient (see~\figref{ref:results:ilp}).}
  \label{ref:method:ilp}
\end{figure}
\subsection{\changed{Constrained Optimization}}
Following~\cite{Buhmann2016}, we represent each candidate microtubule segment $i\in C$ as a node
in a graph with an associated position $p_i=(x_i,y_i,z_i)$.
A priori we do not know which microtubule segments $i \in C$ belong together and form a microtubule. Thus, we connect all microtubule candidates with each other that are below a certain distance threshold $\theta_d$.
More formally, we introduce an undirected graph $G=(V, E)$, where
$V=C\cup\{S\}$ is the set of microtubule candidate segments $C$ augmented with
a special node $S$ and $E\subset V\times V$ is the set of possible links
between them.
The special node $S$ is used to mark the beginning or end of a microtubule
track and is connected to all candidates in $C$.
We further define a set $T=\{(i,j,k)\in V\times C\times V\;|\;(i,j),(j,k)\in E,
i\neq k\}$ of all directly connected triplets on $G$.

As observed in~\cite{Buhmann2016}, we can make use of the fact that
microtubules do not branch and have limited curvature~\cite{gittes1993flexural}.
We encode these priors as constraints and costs respectively, and solve the
resulting optimization problem with an ILP.
%
As outlined in \figref{ref:method:ilp}, and in contrast to~\cite{Buhmann2016},
we formulate consistency and "no-branch" constraints on triplets of connected
nodes $(i,j,k) \in T$ only, leading to an orders of magnitude improvement in
ILP solve time (see~\figref{ref:results:ilp}). To this end, we introduce a
binary indicator variable $I_{i,j,k} \in \{0,1\}$ for each $(i,j,k) \in T$ and
define selection costs $c_{i,j,k}$ for each triplet by propagating costs $c_i$
on nodes and $c_{i,j}$ on edges as follows:
  \begin{equation}
    c_i = \left\{
      \begin{array}{ll}
        \theta_S & \text{if}\;\; i = S\\
        \theta_P & \text{else}
      \end{array}
      \right.
    \;\;\;\;
    \;\;\;\;
    \begin{array}{ll}
      c_{i,j}   &= \theta_D\dist(i,j) + \theta_E\evid(i,j) + c_i + c_j \\
      c_{i,j,k} &= \theta_C\curv(i,j,k) + c_{i,j} + c_{j,k}
    \end{array}
    \text{,}
  \label{equation:cost}
  \end{equation}
  where $\theta_S$ is the cost for beginning/ending a track and $\theta_P < 0$
  is the prior on node selection. $\dist(i,j) = ||p_i - p_j||$ measures the
  distance between candidates $i$ and $j$, whereas $\evid(i, j) = \sum_{p\in
  P_{i,j}} m(p)$ accumulates the
  predicted evidence for microtubules on all voxels on a line $P_{i,j}$
  connecting $i$ and $j$.
  $\curv(i,j,k) = \pi - \angle{(i,j,k)}$ measures deviations of a 180 degree
  angle between two pairs of edges, and thus introduces a cost on
  curvature.
  The values $\theta_S,\theta_P,\theta_D,\theta_E,\theta_C\in\mathbb{R}$ are
  free parameters of the method and found via grid search on a validation
  dataset.

\subsection{\changed{Blockwise Processing}}

%
In order to be able to apply the constraint optimization to arbitrary sized
volumes, we decompose the candidate graph spatially into a set of blocks $B$.
For each block $b\in B$, we define a constant-size context region
$\overline{b}$, which encloses the block and is chosen to be large enough such
that decisions outside the context region are unlikely to change the ILP
solution inside the block.
We next identify sets $S_i \subset B$ of blocks that are pairwise conflict
free, where we define two blocks $a$ and $b$ to be in conflict if $a$ overlaps
with $\overline{b}$.
All blocks of a subset $S_i$ can then be distributed and processed in parallel.
The corresponding ILP for each block $b\in S_i$ is solved within
$\overline{b}$, however, assignments of the binary indicators are only stored
for indicators corresponding to nodes in $b$.
To obtain consistent solutions across block boundaries, existing indicator
assignments from previous runs of conflicting blocks are acknowledged by adding
additional constraints to the block ILP.
See supplement for an illustration.
%
%

\subsection{\changed{Evaluation}}
To evaluate reconstructed tracks against groundtruth, we resample both
reconstruction and groundtruth tracks equidistantly and match nodes based on distance using
Hungarian matching. Results are reported in terms of precision and recall on
edges, which we consider correct if they connect two matched nodes that are
matched to the same track.

\section{Results}
\begin{figure}[t]
    \centering
    \includegraphics[width=1.0\linewidth]{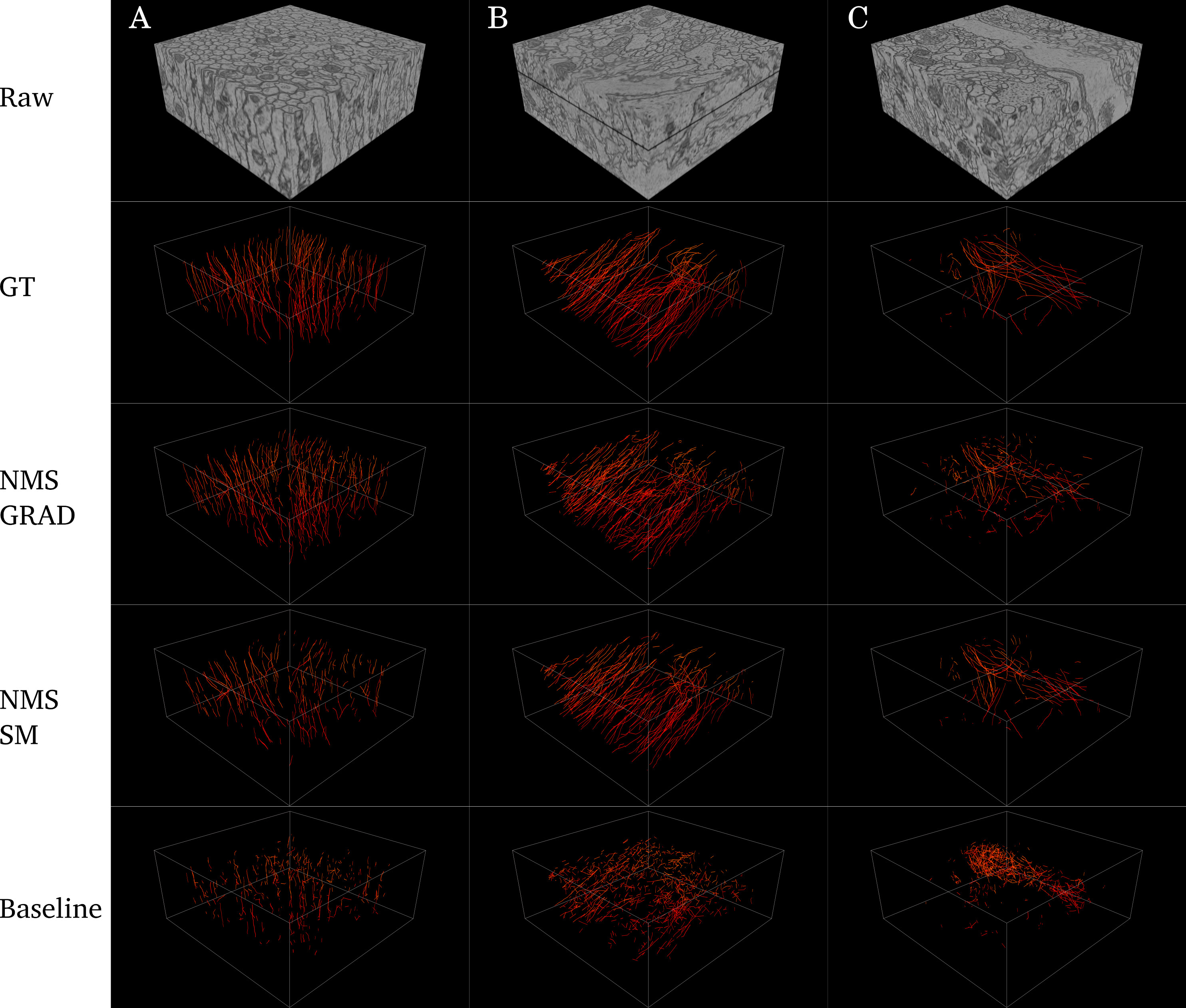}
    \vspace{-5mm}
    \caption{3D rendering of raw EM volumes (Raw), manual tracing (GT) and automatically reconstructed microtubules in CREMI volumes A,B,C for our method (NMS\_GRAD and NMS\_SM) and the considered baseline~\cite{Buhmann2016} using validation best ILP parameters (best viewed on screen).}
    \label{ref:results:3d}
    \vspace{-5mm}
\end{figure}

\begin{figure}[t]
  \pgfplotsset{width=6.6cm,height=5cm}
  \hspace{7mm}
  \begin{subfigure}[t]{0.5\textwidth}
    \begin{tikzpicture}
  \begin{axis}[
      boxplot/draw direction=y,
      ymajorgrids=true,
      xmajorgrids=true,
      ymode=log,
      xlabel=edges,
      ylabel={ILP solve time [s]},
      xticklabel style={font=\tiny,overlay},
      yticklabel style={font=\tiny,overlay},
      ylabel style={font=\tiny,overlay},
      xlabel style={font=\tiny,overlay}]

    \foreach \e in {220, 240, ..., 300}{
      \addplot+[
        color_1,
        filledbox,
        boxplot={draw position=\e,box extend=7}]
          table[y=time] {figures/results/timing/csv_data/G1_\e_ilp_solve_time.csv};
    }

    \foreach \e in {220, 240, ..., 300}{
      \addplot+[
        color_2,
        filledbox,
        boxplot={draw position=\e,box extend=7}]
          table[y=time] {figures/results/timing/csv_data/G2_\e_ilp_solve_time.csv};
    }

  \end{axis}

  \begin{axis}[
      opacity=0,
      xticklabel style={font=\tiny,overlay},
      yticklabel style={font=\tiny,overlay},
      legend style={opacity=1},
      legend pos=north west,
      legend style={font=\tiny},
      legend image post style={scale=2.5},
      legend columns=1]
    \addplot[color_1,filledbox,fill opacity=1,mark=*,only marks] coordinates {(0,0)};
    \addlegendentry{ILP from~\cite{Buhmann2016}}
    \addplot[color_2,filledbox,fill opacity=1,mark=*,only marks] coordinates {(0,0)};
    \addlegendentry{reformulated ILP}
  \end{axis}

\end{tikzpicture}
    \end{subfigure}
  \hfill
  \begin{subfigure}[t]{0.5\textwidth}
    \begin{tikzpicture}
    \begin{axis}[
        ymajorgrids=true,
        xmajorgrids=true,
        legend style={font=\tiny},
        legend image post style={scale=0.5},
        legend columns=1,
        legend pos=north west,
        xlabel=edges,
        ylabel=constraints,
        xticklabel style={font=\tiny,overlay},
        yticklabel style={font=\tiny,overlay},
        ylabel style={font=\tiny,overlay},
        xlabel style={font=\tiny,overlay}]

    \pgfplotstableread{figures/results/timing/csv_data/G1_edges_constraints.csv}\plotdataa
    \pgfplotstableread{figures/results/timing/csv_data/G2_edges_constraints.csv}\plotdatab

    \addplot[color_1,thick,only marks,mark=*]
        table[x=edges,y=constraints] {\plotdataa};

    \addlegendentry{ILP from \cite{Buhmann2016}}

    \addplot[color_2,thick,only marks,mark=*]
        table[x=edges,y=constraints] {\plotdatab};

    \addlegendentry{reformulated ILP}

    \end{axis}
\end{tikzpicture}
  \end{subfigure}
  \vspace{2mm}

  \pgfplotsset{width=6.6cm,height=5.5cm}
  \hspace{7mm}
  \begin{subfigure}{0.5\textwidth}
    \begin{tikzpicture}
    \begin{axis}[
        ymajorgrids=true,
        xmajorgrids=true,
        legend style={font=\tiny},
        legend image post style={scale=0.5},
        legend columns=1,
        legend pos=south west,
        xlabel=recall,
        ylabel=precision,
        xticklabel style={font=\tiny,overlay},
        yticklabel style={font=\tiny,overlay},
        ylabel style={font=\tiny,overlay},
        xlabel style={font=\tiny,overlay},
        xmin=0,
        xmax=1,
        ymin=0,
        ymax=1
    ]

    \pgfplotstableread{figures/results/method_comparison/baseline_test.csv}\plotdataaunsorted
    \pgfplotstableread{figures/results/method_comparison/softmask_test.csv}\plotdatabunsorted
    \pgfplotstableread{figures/results/method_comparison/lsd_test.csv}\plotdatacunsorted
    \pgfplotstableread{figures/results/method_comparison/cc_test.csv}\plotdatadunsorted
    \pgfplotstableread{figures/results/method_comparison/rfc_test.csv}\plotdataeunsorted

    \pgfplotstablesort{\plotdataa}{\plotdataaunsorted}
    \pgfplotstablesort{\plotdatab}{\plotdatabunsorted}
    \pgfplotstablesort{\plotdatac}{\plotdatacunsorted}
    \pgfplotstablesort{\plotdatad}{\plotdatadunsorted}
    \pgfplotstablesort{\plotdatae}{\plotdataeunsorted}

    \addplot[color_1,thick]
        table[x=recall,y=precision] {\plotdataa};

    \addlegendentry{baseline}

    \addplot[color_2,thick]
        table[x=recall,y=precision] {\plotdatab};

    \addlegendentry{NMS\_SM}

    \addplot[color_3,thick]
        table[x=recall,y=precision] {\plotdatac};
    \addlegendentry{NMS\_GRAD}

    \addplot[color_4,thick]
        table[x=recall,y=precision] {\plotdatad};
    \addlegendentry{CC\_GRAD}

    \addplot[color_5,thick]
        table[x=recall,y=precision] {\plotdatae};
    \addlegendentry{NMS\_RFC}

    \addplot[
      color_1!60!black,
      thick,
      mark=o,
      mark size=2,
      x filter/.code={
        \pgfplotstablegetelem{\coordindex}{validation_best}\of{\plotdataa}
        \ifnum\pgfplotsretval=1
        \else
        \def\pgfmathresult{}
        \fi
        }]
        table[x=recall,y=precision] {\plotdataa};
    \addplot[
      color_1!60!black,
      thick,
      mark=star,
      mark size=2,
      x filter/.code={
        \pgfplotstablegetelem{\coordindex}{test_best}\of{\plotdataa}
        \ifnum\pgfplotsretval=1
        \else
        \def\pgfmathresult{}
        \fi
        }]
        table[x=recall,y=precision] {\plotdataa};

    \addplot[
      color_2!60!black,
      thick,
      mark=o,
      mark size=2,
      x filter/.code={
        \pgfplotstablegetelem{\coordindex}{validation_best}\of{\plotdatab}
        \ifnum\pgfplotsretval=1
        \else
        \def\pgfmathresult{}
        \fi
        }]
        table[x=recall,y=precision] {\plotdatab};
    \addplot[
      color_2!60!black,
      thick,
      mark=star,
      mark size=2,
      x filter/.code={
        \pgfplotstablegetelem{\coordindex}{test_best}\of{\plotdatab}
        \ifnum\pgfplotsretval=1
        \else
        \def\pgfmathresult{}
        \fi
        }]
        table[x=recall,y=precision] {\plotdatab};

    \addplot[
      color_3!60!black,
      thick,
      mark=o,
      mark size=2,
      x filter/.code={
        \pgfplotstablegetelem{\coordindex}{validation_best}\of{\plotdatac}
        \ifnum\pgfplotsretval=1
        \else
        \def\pgfmathresult{}
        \fi
        }]
        table[x=recall,y=precision] {\plotdatac};
    \addplot[
      color_3!60!black,
      thick,
      mark=star,
      mark size=2,
      x filter/.code={
        \pgfplotstablegetelem{\coordindex}{test_best}\of{\plotdatac}
        \ifnum\pgfplotsretval=1
        \else
        \def\pgfmathresult{}
        \fi
        }]
        table[x=recall,y=precision] {\plotdatac};

    \addplot[
      color_4!60!black,
      thick,
      mark=o,
      mark size=2,
      x filter/.code={
        \pgfplotstablegetelem{\coordindex}{validation_best}\of{\plotdatad}
        \ifnum\pgfplotsretval=1
        \else
        \def\pgfmathresult{}
        \fi
        }]
        table[x=recall,y=precision] {\plotdatad};
    \addplot[
      color_4!60!black,
      thick,
      mark=star,
      mark size=2,
      x filter/.code={
        \pgfplotstablegetelem{\coordindex}{test_best}\of{\plotdatad}
        \ifnum\pgfplotsretval=1
        \else
        \def\pgfmathresult{}
        \fi
        }]
        table[x=recall,y=precision] {\plotdatad};

    \addplot[
      color_5!60!black,
      thick,
      mark=o,
      mark size=2,
      x filter/.code={
        \pgfplotstablegetelem{\coordindex}{validation_best}\of{\plotdatae}
        \ifnum\pgfplotsretval=1
        \else
        \def\pgfmathresult{}
        \fi
        }]
        table[x=recall,y=precision] {\plotdatae};
    \addplot[
      color_5!60!black,
      thick,
      mark=star,
      mark size=2,
      x filter/.code={
        \pgfplotstablegetelem{\coordindex}{test_best}\of{\plotdatae}
        \ifnum\pgfplotsretval=1
        \else
        \def\pgfmathresult{}
        \fi
        }]
        table[x=recall,y=precision] {\plotdatae};

    \end{axis}
\end{tikzpicture}
  \end{subfigure}
  \hfill
  \begin{subfigure}{0.5\textwidth}
    \hspace{1mm}
    \includegraphics[width=4.5cm]{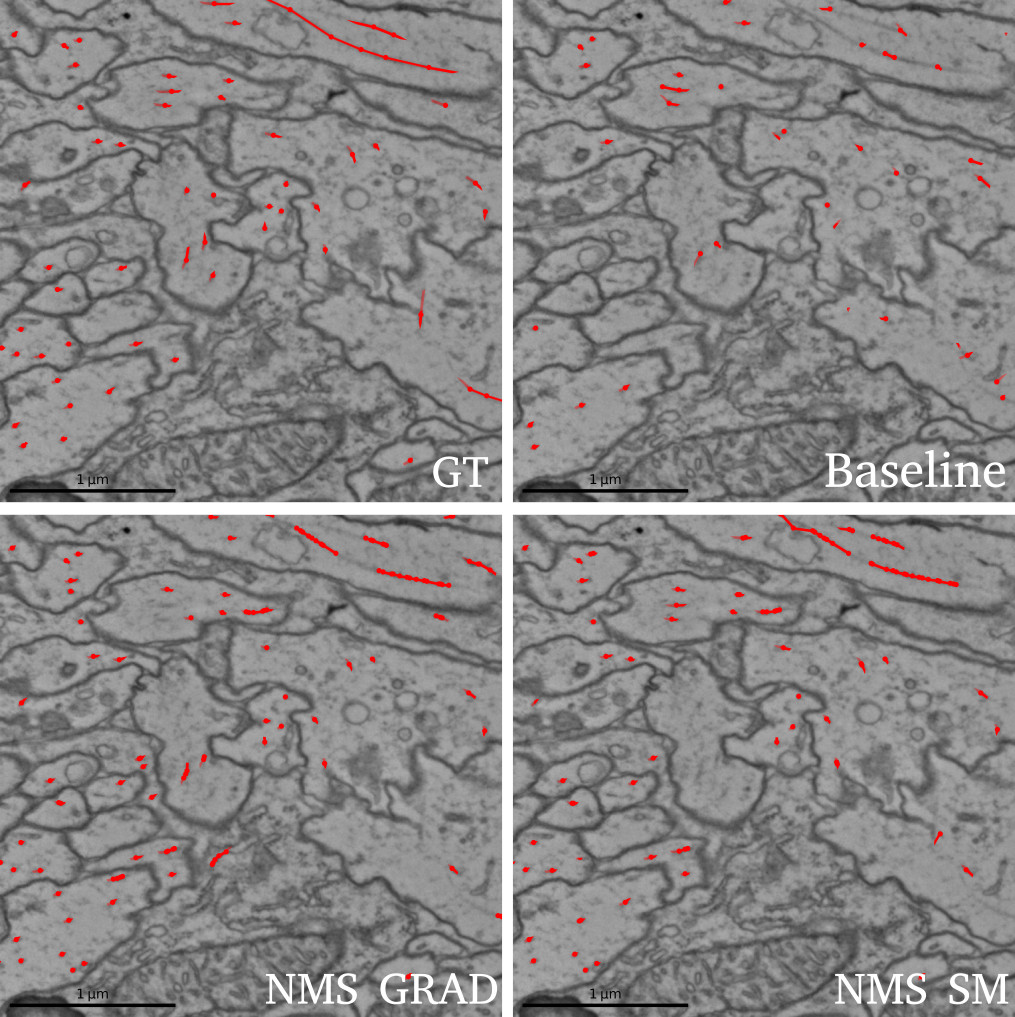}
  \end{subfigure}
  \vspace{-2mm}
  \caption{
    {\bf Top row:} comparison of ILP formulations on random candidate graphs in
    terms of solve time (left) and number of constraints (right). Solve times
    have been obtained from 54 different ILP parameter configurations
    $\theta_{S,P,D,E,C}$ on an Intel Xeon(R), 2.40GHz x 16 CPU processor using
    the Gurobi optimizer.
    {\bf \changed{Bottom row, left}:} Comparison of our method (NMS\_SM and NMS\_GRAD) to
    the baseline~\cite{Buhmann2016} and two ablation experiments CC\_GRAD 
    (NMS replaced with connected component candidate extraction) and 
    NMS\_RFC (UNet replaced with RFC). 
    Shown are precision and recall for varying
    values of the start/end edge prior $\theta_{S}$ averaged over the test
    datasets A,B,C. The validation and test best are highlighted with circles
    and stars, respectively.
    {\bf Bottom row, right:} Qualitative results on sample B (best viewed on
    screen).
  }
  \label{ref:results:ilp}
\end{figure}
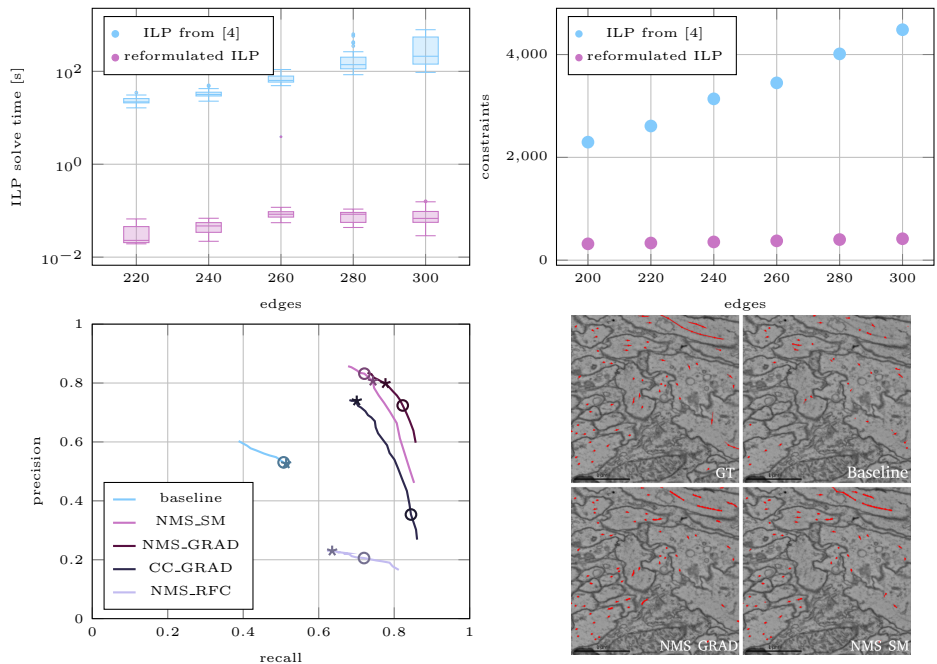

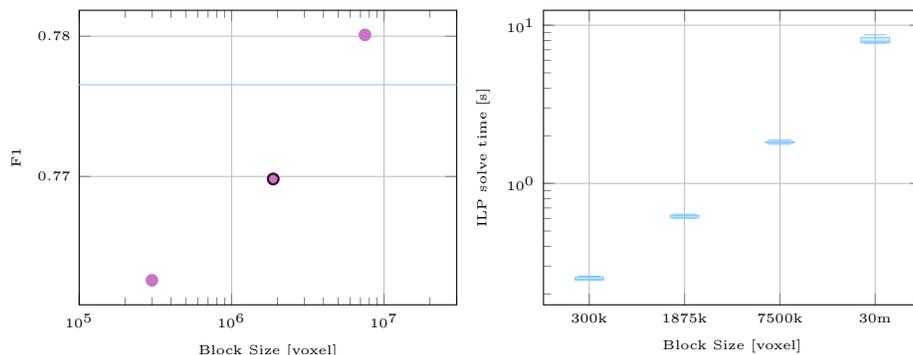
\begin{figure}[t]
\pgfplotsset{width=6.6cm,height=5.5cm}
    \hspace{7mm}
  \begin{subfigure}{0.5\textwidth}
    \begin{tikzpicture}
    \begin{axis}[
        ymajorgrids=true,
        xmode=log,
        xmajorgrids=true,
        legend style={font=\tiny},
        legend image post style={scale=0.5},
        legend columns=1,
        legend pos=north west,
        xmin=100000,
        xmax=30000000,
        xlabel={Block Size [voxel]},
        ylabel=F1,
        xticklabel style={font=\tiny,overlay},
        yticklabel style={font=\tiny,overlay},
        ylabel style={font=\tiny,overlay},
        xlabel style={font=\tiny,overlay}]

    \pgfplotstableread{figures/results/block/block_size.csv}\plotdataa

    \addplot[color_2,thick,only marks,mark=*]
        table[x=Block_Size,y=F1] {\plotdataa};

    \addplot[
      color_3!60!black,
      thick,
      mark=o,
      mark size=2,
      x filter/.code={
        \pgfplotstablegetelem{\coordindex}{validation}\of{\plotdataa}
        \ifnum\pgfplotsretval=1
        \else
        \def\pgfmathresult{}
        \fi
        }]
        table[x=Block_Size,y=F1] {\plotdataa};

    \addplot[color_1,domain=10000:30000000] {0.77327};

    \end{axis}
\end{tikzpicture}
  \end{subfigure}
  \begin{subfigure}{0.5\textwidth}
    \begin{tikzpicture}
  \begin{axis}[
      boxplot/draw direction=y,
      ymajorgrids=true,
      xmajorgrids=true,
      ymode=log,
      xlabel={Block Size [voxel]},
      ylabel={ILP solve time [s]},
      xtick={1,2,3,4},
      xticklabels={300k, 1875k, 7500k, 30m},
      xticklabel style={font=\tiny,overlay},
      yticklabel style={font=\tiny,overlay},
      ylabel style={font=\tiny,overlay},
      xlabel style={font=\tiny,overlay}]

    \foreach \e in {300000, 1875000,7500000,30000000}{
      \addplot+[
        color_1,
        filledbox,
        boxplot={box extend=0.3}]
          table[y=time] {figures/results/block/timing_run\e.csv};
    }

  \end{axis}
\end{tikzpicture}
  \end{subfigure}
  \caption{\textbf{Left}: Accuracy as a function of block size over several
  orders of magnitude. Shown are the F1 scores, averaged over test data sets A,
  B and C, using validation best parameters NMS\_GRAD. Interestingly, for some
  sizes, solving the ILP block-wise results in higher F1 scores than solving
  the ILP to global optimality (blue line). However, it should be noted that
  the differences in F1 score are minor and likely not significant. The black
  circle indicates the block size we used for all reported results.
  \textbf{Right}: Box plot of ILP solve time per block as a function of block size. Shown is the wall-clock time
  needed to solve the ILP for one block, measured for ten runs, on test cube B using
    validation best parameters NMS\_GRAD. \changed{Note that in contrast to accuracy, solve time is strongly affected by block size. This implies that we are able to process large volumes by solving the ILP in a blockwise manner, without a significant decrease in accuracy.}}
  \label{ref:results:blocks}
\end{figure}


\subsection{\changed{Dataset}}
We densely annotated microtubules in eight 1.2x4x4$\mu m$ (30x1000x1000 voxel)
volumes of EM data in all six CREMI~\footnote{MICCAI Challenge on Circuit
Reconstruction in EM Images, \url{https://cremi.org}.} volumes A, B, C, A+, B+,
C+ using Knossos~\cite{boergens2017webknossos} and split the data in training
(A+, B+, C+), validation ($\text{B+}_{v}$, $\text{B}_{v}$) and test (A, B, C)
sets.
\subsection{\changed{Comparison}}
\textit{NMS\_*} models refer to the model described in the methods section, where NMS\_SM uses a 3D UNet predicting microtubule score only, NMS\_\-GRAD additionally predicts spatial gradients of the microtubule score up to second order and NMS\_\-RFC uses a random forest classifier (RFC) instead of a 3D UNet.
For each, we first select the best performing UNet architecture (for NMS\_\-RFC we interactively train an RFC using Ilastik~\cite{sommer2011ilastik}) and NMS
candidate extraction threshold in terms of recovered candidates on the
validation datasets, followed by a grid search over the distance threshold
$\theta_d$ and ILP parameters for 150 different parameter combinations. 
For the NMS candidate extraction we use a window size of 1x10x10 voxels for the
first NMS pass to offset the anisotropic resolution of 40x4x4~nm. 
For the second NMS pass we use a window size of 1x3x3~voxels, removing double
detections.\\ \\
\textit{Baseline} refers to an adaptation\footnote{We use our ILP formulation,
which was necessary to process larger volumes.} of the method
in~\cite{Buhmann2016}, that uses an RFC for
prediction, z section-wise connected component (CC) analysis on the thresholded
microtubule scores for candidate extraction, and a fixed orientation estimate
for each microtubule candidate pointing in the z direction\footnote{Orientation
estimate used in~\cite{Buhmann2016} (direct communication with authors).}.
For the baseline we interactively train two (for microtubules of
different angles of incidence on the imaging plane) RFCs on training volumes A,
B, C using Ilastik~\cite{sommer2011ilastik}. We find the threshold for CC
candidate extraction, distance threshold $\theta_d$ and ILP parameters via grid
search over 242 parameter configurations on the validation set. For an overview
see Table~\ref{table:results:models}.
%
\subsection{\changed{Test Results}}
\figref{ref:results:ilp} shows that both variants of our proposed model
outperform the prior state of the art~\cite{Buhmann2016} substantially.
Averaged over test data sets A,B,C, we demonstrate a 53\% increase in accuracy
for NMS\_GRAD.
Table~\ref{table:results:models} further shows test best F1 scores for each
individual dataset.
In accordance with the qualitative results shown in \figref{ref:results:3d},
NMS\_GRAD performs substantially better for test set A while NMS\_SM is more
accurate for volumes B and C.
Ablation experiments show that CC candidate extraction leads to overall less accurate reconstructions. 
Exchanging the UNet with an RFC while retaining NMS candidate extraction seriously harms performance, resulting in large numbers of false positive detections. 
For extended qualitative results, including reconstruction of microtubules in the Calyx, a 76 x 52 x 64 $\mu m$ region of the \textit{Drosophila Melanogaster} brain, see supplement.
\begin{table}[t]
\caption{Model overview and test best F1 score by data set.}
\label{table:results:models}
\centering 
\begin{tabular}{c c c c c c c c}
\hline
    Model & Prediction & Cand. Extr. & Edge Score & A & B & C & \changed{Avg}\\
\hline
    NMS\_GRAD & UNet+GRAD & NMS & Evidence & \textbf{0.784} & 0.827 & 0.757 & \textbf{0.789}\\
    NMS\_SM & UNet & NMS & Evidence & 0.711 & \textbf{0.828} & \textbf{0.785} & 0.775\\
    Baseline & RFC & CC  & Orientation & 0.454 & 0.547 & 0.549 & 0.517\\
    CC\_GRAD & \changed{UNet}+GRAD & CC & Evidence & 0.660 & 0.723 & 0.537 & 0.640\\
    NMS\_RFC & RFC & NMS & Evidence & 0.366 & 0.375 & 0.302 & 0.348\\
\hline
\end{tabular}
\end{table}

\section{Discussion}

%

%
%
Although some of our improvements in accuracy can be attributed to the use of a
deep learning classifier, the presented method relies mostly on an effective
way of incorporating biological priors in the form of constraint optimization.
In particular our ablation studies (CC\_GRAD) show that the strided NMS-based candidate extraction method 
positively impacts accuracy: Since a single microtubule could potentially 
extend far in the x-y imaging plane, it is
not sufficient to represent candidates in one plane by a single node, as done
in~\cite{Buhmann2016}. The strided NMS detections homogenize the candidate
graph and likely allow transferring our method to datasets of different
resolutions.
A potential downside is poor precision when combined with extremely noisy microtubule score predictions $m$ (see NMS\_RFC). 
In this case NMS on a grid extracts too
many candidate segments, and besides structural priors, the only remaining cost we use to extract 
final microtubule tracks is directly derived from the (noisy) predicted microtubule score $m$ (see equation \eqref{equation:cost}).
Note that the baseline does not suffer as much from noisy microtubule scores, because it uses a 
fixed orientation prior and is thus limited to a subset of microtubules in any given volume. 
%
%
%
%
%
Finally, the reformulation of the ILP and the block-wise processing scheme
result in a dramatic speed-up and the ability to perform distributed, consistent
tracking, which is required to process petabyte-sized datasets.

\section{Acknowledgements}
We thank Tri Nguyen and Caroline Malin-Mayor for code contribution; Arlo Sheridan for helpful discussions and Albert Cardona for his contagious enthusiasm and support. This work was supported by Howard Hughes Medical Institute and Swiss National Science Foundation (SNF grant 205321L 160133).

\bibliographystyle{splncs04}
\bibliography{paper1746}

\begin{thebibliography}{10}
\providecommand{\url}[1]{\texttt{#1}}
\providecommand{\urlprefix}{URL }
\providecommand{\doi}[1]{https://doi.org/#1}

\bibitem{boergens2017webknossos}
Boergens, K.M., Berning, M., Bocklisch, T., Br{\"a}unlein, D., Drawitsch, F.,
  Frohnhofen, J., Herold, T., Otto, P., Rzepka, N., Werkmeister, T., et~al.:
  webknossos: efficient online 3d data annotation for connectomics. nature
  methods  \textbf{14}(7),  691--694 (2017)

\bibitem{synapse_julia}
Buhmann, J., Krause, R., Ceballos~Lentini, R., Eckstein, N., Cook, M., Turaga,
  S., Funke, J.: Synaptic partner prediction from point annotations in insect
  brains. Medical Image Computing and Computer Assisted Intervention – MICCAI
  2018  \textbf{11071} (09 2018)

\bibitem{Buhmann2019}
Buhmann, J., Sheridan, A., Gerhard, S., Krause, R., Nguyen, T., Heinrich, L.,
  Schlegel, P., Lee, W.C.A., Wilson, R., Saalfeld, S., Jefferis, G., Bock, D.,
  Turaga, S., Cook, M., Funke, J.: Automatic detection of synaptic partners in
  a whole-brain drosophila em dataset. bioRxiv  (2019).
  \doi{10.1101/2019.12.12.874172},
  \url{https://www.biorxiv.org/content/early/2019/12/13/2019.12.12.874172}

\bibitem{Buhmann2016}
Buhmann, J.M., Gerhard, S., Cook, M., Funke, J.: Tracking of microtubules in
  anisotropic volumes of neural tissue. In: 2016 {IEEE} 13th International
  Symposium on Biomedical Imaging ({ISBI}) (2016).
  \doi{10.1109/isbi.2016.7493275},
  \url{http://dx.doi.org/10.1109/ISBI.2016.7493275}

\bibitem{mitos2}
Cheng, H.C., Varshney, A.: Volume segmentation using convolutional neural
  networks with limited training data. In: 2017 IEEE international conference
  on image processing (ICIP). pp. 590--594. IEEE (2017)

\bibitem{mitos3}
Dorkenwald, S., Schubert, P.J., Killinger, M.F., Urban, G., Mikula, S., Svara,
  F., Kornfeld, J.: Automated synaptic connectivity inference for volume
  electron microscopy. Nature methods  \textbf{14}(4),  435--442 (2017)

\bibitem{mala}
{Funke}, J., {Tschopp}, F.D., {Grisaitis}, W., {Sheridan}, A., {Singh}, C.,
  {Saalfeld}, S., {Turaga}, S.C.: Large scale image segmentation with
  structured loss based deep learning for connectome reconstruction. IEEE
  Transactions on Pattern Analysis and Machine Intelligence pp.~1--1 (2018).
  \doi{10.1109/TPAMI.2018.2835450}

\bibitem{gittes1993flexural}
Gittes, F., Mickey, B., Nettleton, J., Howard, J.: Flexural rigidity of
  microtubules and actin filaments measured from thermal fluctuations in shape.
  The Journal of cell biology  \textbf{120}(4),  923--934 (1993)

\bibitem{heinrich2018synaptic}
Heinrich, L., Funke, J., Pape, C., Nunez-Iglesias, J., Saalfeld, S.: Synaptic
  cleft segmentation in non-isotropic volume electron microscopy of the
  complete drosophila brain. In: International Conference on Medical Image
  Computing and Computer-Assisted Intervention. pp. 317--325. Springer (2018)

\bibitem{huang2018fully}
Huang, G.B., Scheffer, L.K., Plaza, S.M.: Fully-automatic synapse prediction
  and validation on a large data set. Frontiers in neural circuits
  \textbf{12}, ~87 (2018)

\bibitem{januszewski2018high}
Januszewski, M., Kornfeld, J., Li, P.H., Pope, A., Blakely, T., Lindsey, L.,
  Maitin-Shepard, J., Tyka, M., Denk, W., Jain, V.: High-precision automated
  reconstruction of neurons with flood-filling networks. Nature methods p.~1
  (2018)

\bibitem{synapses_02}
Kreshuk, A., Funke, J., Cardona, A., Hamprecht, F.A.: Who is talking to whom:
  Synaptic partner detection in anisotropic volumes of insect brain. In: Navab,
  N., Hornegger, J., Wells, W.M., Frangi, A. (eds.) Medical Image Computing and
  Computer-Assisted Intervention -- MICCAI 2015. pp. 661--668. Springer
  International Publishing, Cham (2015)

\bibitem{Lee2019}
{Lee}, K., {Lu}, R., {Luther}, K., {Seung}, H.S.: {Learning Dense Voxel
  Embeddings for 3D Neuron Reconstruction}. arXiv e-prints arXiv:1909.09872
  (Sep 2019)

\bibitem{Lee2017superhuman}
Lee, K., Zung, J., Li, P., Jain, V., Seung, H.S.: Superhuman accuracy on the
  snemi3d connectomics challenge. arXiv preprint arXiv:1706.00120  (2017)

\bibitem{mt_review_0}
Nogales, E.: Structural insights into microtubule function. Annual Review of
  Biochemistry  \textbf{69}(1),  277--302 (2000)

\bibitem{ronneberger2015u}
Ronneberger, O., Fischer, P., Brox, T.: U-net: Convolutional networks for
  biomedical image segmentation. In: International Conference on Medical image
  computing and computer-assisted intervention. pp. 234--241. Springer (2015)

\bibitem{schneider2016quantitative}
Schneider-Mizell, C.M., Gerhard, S., Longair, M., Kazimiers, T., Li, F., Zwart,
  M.F., Champion, A., Midgley, F.M., Fetter, R.D., Saalfeld, S., et~al.:
  Quantitative neuroanatomy for connectomics in drosophila. eLife  \textbf{5},
  e12059 (2016)

\bibitem{sommer2011ilastik}
Sommer, C., Straehle, C.N., Koethe, U., Hamprecht, F.A., et~al.: Ilastik:
  Interactive learning and segmentation toolkit. In: ISBI. vol.~2, p.~8 (2011)

\bibitem{synapses_03}
Staffler, B., Berning, M., Boergens, K.M., Gour, A., van~der Smagt, P.,
  Helmstaedter, M.: Synem, automated synapse detection for connectomics. Elife
  \textbf{6},  e26414 (2017)

\bibitem{FIB25}
Takemura, S.y., Xu, C.S., Lu, Z., Rivlin, P.K., Parag, T., Olbris, D.J., Plaza,
  S., Zhao, T., Katz, W.T., Umayam, L., et~al.: Synaptic circuits and their
  variations within different columns in the visual system of drosophila.
  Proceedings of the National Academy of Sciences  \textbf{112}(44),
  13711--13716 (2015)

\bibitem{mitos}
Xiao, C., Chen, X., Li, W., Li, L., Wang, L., Xie, Q., Han, H.: Automatic
  mitochondria segmentation for em data using a 3d supervised convolutional
  network. Frontiers in Neuroanatomy  \textbf{12}, ~92 (2018).
  \doi{10.3389/fnana.2018.00092},
  \url{https://www.frontiersin.org/article/10.3389/fnana.2018.00092}

\bibitem{zheng2018complete}
Zheng, Z., Lauritzen, J.S., Perlman, E., Robinson, C.G., Nichols, M., Milkie,
  D., Torrens, O., Price, J., Fisher, C.B., Sharifi, N., et~al.: A complete
  electron microscopy volume of the brain of adult drosophila melanogaster.
  Cell  \textbf{174}(3),  730--743 (2018)

\end{thebibliography}
\end{document}